%% file: ms.tex
\documentclass[letterpaper]{article}
\usepackage{aaai18}
\usepackage{times}
\usepackage{helvet}
\usepackage{courier}
\usepackage{amsmath}
\usepackage{booktabs}
\usepackage[usenames, dvipsnames]{color}
\usepackage{color}
\usepackage{hyperref}
\usepackage{xargs}                      % Use more than one optional parameter in a new commands
\usepackage{verbatim}
\usepackage{caption}
\usepackage{subcaption}
\usepackage{graphicx}

\usepackage{amssymb}
\usepackage{physics}

\begin{document}

\title{Improving Language Modelling with Noise Contrastive Estimation}
\author{Farhana Ferdousi Liza \and Marek Grzes\\
University of Kent\\
Canterbury, CT2 7NZ, UK\\
{\{fl207, m.grzes\}}@kent.ac.uk}

%\nocopyrightcommand
\copyrighttext{}

\maketitle

\begin{abstract}
\begin{quote}
\input{abstract}
\end{quote}
\end{abstract}

\section{Introduction}
\input{introduction}

%moved from here
\section{Background}
\label{sec:back}
%\input{model}
\input{background}
\section{Our Approach}
\label{sec:model}
\input{nce_training}

\input{table_compare_state_of_the_art}

\section{Experimental Methodology and Implementation}
\label{sec:exp}
\input{experiment}

\section{Results and discussion}
\label{sec:res}
\input{results}
\input{result_discussion}
\section{Conclusion}
\input{conclusion}

%\section{Appendix}
%\input{appendix}
%\bibliography{../bibliography_deep_nlp}
\bibliography{ms}
\bibliographystyle{aaai}
\end{document}

%% file: abstract.tex
Neural language models do not scale well when the vocabulary is large. Noise contrastive estimation (NCE) is a sampling-based method that allows for fast learning with large vocabularies. Although NCE has shown promising performance in neural machine translation, it was considered to be an unsuccessful approach for language modelling. A sufficient investigation of the hyperparameters in the NCE-based neural language models was also missing. In this paper, we showed that NCE can be a successful approach in neural language modelling when the hyperparameters of a neural network are tuned appropriately. We introduced the `search-then-converge' learning rate schedule for NCE and designed a heuristic that specifies how to use this schedule. The impact of the other important hyperparameters, such as the dropout rate and the weight initialisation range, was also demonstrated. We showed that appropriate tuning of NCE-based neural language models outperforms the state-of-the-art single-model methods on a popular benchmark. 

% state-of-the-art results and did the best among the other single-model approach available in the literature.

% The hyperparameters have different inpact tha MLE counterpart. Particularly,

% by converting the original estimation problem into a nonlinear logistic regression problem. 

%% file: introduction.tex
Statistical language models (LMs), which predict the probability of a next word given its context, play an important role in many downstream applications such as machine translation, question answering and text summarisation. Neural language models which apply various architectures  \cite{bengio2003neural,conf/interspeech/MikolovKBCK10,DBLP:journals/corr/JozefowiczVSSW16} have recently demonstrated significant achievements.
%chelba2013one

In real applications, the vocabulary size is large and the language models have to estimate a probability distribution over many words. The need for a normalised probability distribution becomes a computational bottleneck because the normalisation constant (i.e. the partition function) has to be computed for the output layer.

%Language model: In a neural language model the conditional distribution corresponding to context $h$, $P^{h}(w)$,is defined as

%$${P_{\theta}}^{h} = \frac{exp(S_{\theta}(w,h))}{\sum_{w' \in V}{exp(S_{\theta}(w',h))}}$$

%For training language model the partition function (to normalize the model) needs to be calculated, which involves the calculation of a compatibility score ($S_{\theta}(w,h)$) over all words in the vocabulary. If we have a large vocabulary then this makes the whole training very inefficient computationally. 

%morin2005hierarchical removed to save space
Many solutions have been proposed to address the computational complexity of the partition function. Several approaches try to make it more efficient, e.g., hierarchical softmax \cite{MnihHinton2009}, %5947610
a shortlisting method \cite{Schwenk2007492}, or self-normalisation techniques \cite{DBLP:journals/corr/ChenGA15}. %devlin2014fast, Andreas:2015:ASL:2969239.2969438
The other methods, such as importance sampling \cite{Bengio+Senecal-2003,DBLP:journals/corr/JeanCMB14} %bengio2008adaptive
or noise contrastive estimation \cite{Mnih12afast}, aim at computing an unnormalised statistical model.
%\todo[inline]{what is a theoretically unnormalised statistical model? Please check the paragraph above. I am not sure if my corrections make technical sense. Ans: IS and NCE are used to learn unnormalized model. A theoretical unnormalized statistical model is one where density function does not integrate to one. MG: I removed the word "theoretically".}

In this paper, we investigate noise contrastive estimation (NCE) because of its statistical consistency, and the fact that its potential has not been sufficiently explored in the literature on language models (LMs). NCE has also achieved promising results in machine translation \cite{conf/emnlp/VaswaniZFC13,conf/naacl/BaltescuB15} which indicates that its performance on language models could be better than what is known in current research. %Incorporating traditional LMs into real-life applications, such as machine translation, is challenging because the partition function in LMs is a bottleneck. %NCE solves the scaling problem (see Figure 2 of \cite{conf/emnlp/VaswaniZFC13}), where it was shown that training time does not increase for increasing vocabulary size for a fixed number of noise sample.
%NCE led to promising results in machine translation \cite{conf/emnlp/VaswaniZFC13,conf/naacl/BaltescuB15} but the potential of NCE has not been shown for language models. 

NCE was first proposed in \cite{gutmann2010noise} as an estimation principle for unnormalised statistical models. Unnormalised statistical models compute values which, in contrast to formal probabilities, do not add up to one. In order to normalise those values so that they become valid probabilities, they can be divided by the partition function. However, the partition function is computationally expensive to compute when the number of outcomes is large. Therefore, instead of calculating the partition function, NCE converts the original estimation problem into a nonlinear logistic regression problem which discriminates the noise samples generated from a known (noise) distribution from the original data samples. NCE is statistically consistent and more stable than other Monte Carlo methods such as importance sampling \cite{Mnih12afast}. In \cite{gutmann2010noise}, NCE achieved the best trade-off between computational and statistical efficiency when compared against importance sampling, contrastive divergence \cite{hinton2002training}, and score matching \cite{hyvarinen2005estimation}. This method has also been applied in language modelling and machine translation \cite{Mnih12afast,conf/emnlp/VaswaniZFC13,conf/naacl/BaltescuB15,Zoph2016SimpleFN}. 

Many features of NCE are not understood, especially the hyperparameters in deep learning when NCE is used at the output layer. Also, comparisons against single-model softmax have never shown that NCE can compete with softmax on those tasks on which softmax is feasible. % Using a standard and well-studied benchmark, in this paper, we show, for the first time, that NCE can outperform softmax in terms of quality.

Our results are clearly surprising in the face of the existing literature which generally indicates that NCE is an inferior method. % is challenging as we have to train the model with less amount of data and deal with overfitting issue.%(log-linear model LM), \cite{Zoph2016SimpleFN} (Neural LM and MT), \cite{conf/emnlp/VaswaniZFC13} (Neural LM and MT), \cite{conf/naacl/BaltescuB15} (Neural LM and MT). \info[inline]{Seems like NCE is better for LM when it can further applicable to improve MT} %\info{  Recently there has been a trend to improve language models by us ing ever larger amounts of training data, for instance several billions of words of English news paper texts.  However, such enormous amounts of data are not necessarily available for many European languages, and we need to deploy methods that take the best advantage of a limited quantity of training data. So we should have a good LM with low number of training data.}
% http://www.statmt.org/mtm2/data/Holger_Schwenk_proposal.pdf
For example, studying language modelling in \cite{DBLP:journals/corr/JozefowiczVSSW16}, the authors argued that importance sampling (IS) may be better than NCE as IS optimises a multiclass classification task whereas NCE is solving a binary task. %The argument is based on the fact that updates to the logits in IS are tied whereas in NCE they are independent.
%\todo[inline]{what does it mean? maybe we can remove this sentence with logits?Ans: Ok, This line is not important.} 
In \cite{DBLP:journals/corr/JozefowiczVSSW16}, the authors managed to improve the results on language modelling using IS, whereas similar improvements for NCE were not found. Overall, the current literature does not have substantial, empirical evidence that NCE is a powerful method. %The success of IS was surprising because in practice the high variance of the IS estimates makes learning unstable. 

Another example which demonstrates weak performance of NCE on language modelling is \cite{DBLP:journals/corr/ChenGA15}. The authors explain that a limited number (50) of noise samples in NCE does not allow for frequent sampling of every word in a large vocabulary. The number of noise samples has to be relatively small to make the method feasible. %They used the unigram distribution in their NCE experiments following \cite{Mnih12afast}.
In their experiments, NCE performed better than softmax only on billionW, a dataset on which softmax is very slow due to a very large vocabulary. So NCE was better only because softmax was not feasible on a large vocabulary. In this paper, we show, for the first time, that NCE can outperform softmax in a situation when softmax is feasible and it is known to perform very well. To demonstrate that, we used the Penn Tree Bank (PTB) dataset\footnote{http://www.fit.vutbr.cz/~imikolov/rnnlm/simple-examples.tgz} \cite{marcus1993building}, which is a popular language modelling benchmark with a vocabulary size of 10k words. Softmax is known for competitive performance on this data, and it is feasible to apply it to this data using GPUs.

%%%%%%%%%%%%%
%however we get very good result (ppl 72.767) with smaller PTB dataset with log-uniform distribution which is better than the best softmax result of ppl 78.4.The noise sample size is certainly important, as we get the ppl of  69.975 with sample size 600. 
%%%%%%%%%%%%%

% If maximum likelihood parameters can be computed, they can be seen as the best parameters for a model. They can, however, only be obtained if an analytical expression for the partition function is available, which is hard to compute in language modelling. Also, the sufficient statistic in this case is the entire dataset, i.e., there is no small, sufficient statistic that summaries the data. \todo[inline]{Could I read about this somewhere?Ans: Sorry, I dont understand this question clearly. But you can read the following two links.}
% MG: I think that we should remove the paragraph above.
% My intuition was as followes, as sufficient statistic has to have all the information needed to compute any estimate of the parameter, for LM, first of all, it is impossible to say there exist a dataset which can give all information needed to learn the LM parameters, because, language is very creative, thus sparse and there are rare/out of voc words. Following can be read: Please see Proposition 5 at https://courses.cit.cornell.edu/econ620/reviewm5.pdf and 4.2.3 at http://www.maths.manchester.ac.uk/~peterf/CSI_ch4_part1.pdf.
There exist papers in which the researchers tried to create conditions which make softmax feasible to be executed on large datasets. For example, the experiments in \cite{conf/naacl/BaltescuB15} are based on a few billions of training examples and a vocabulary with over 100k tokens. To manage the softmax computation, the authors partitioned the vocabulary into $k$ classes. Under those conditions, the authors showed that NCE performed almost as well as softmax. Softmax in their comparisons was approximate, however, due to partitioning. In our paper, the goal is to compete with the original softmax without any approximations. % maximum likelihood training with stochastic gradient descent, which is aligned with the NCE theory. Theoretically, when the noise sample is sufficiently large, the NCE gradient approaches the MLE gradient.
%(see the nce derivation at appendix for clarity \ref{SS:MLE_NCE_GRADIENT}). 
%Although in \cite{DBLP:journals/corr/SaxeMG13,DBLP:journals/corr/DauphinPGCGB14,DBLP:journals/corr/ChoromanskaHMAL14}, authors said that local minima are not a significant problem for training large neural networks.
Our aims are justified by the following reasoning. In theory, NCE, being a statistically consistent method, converges to the maximum likelihood estimation method when the number of noise samples is increased. However, the fact that NCE solves a different optimisation problem means that stochastic gradient descent applied to neural networks with NCE may find a different, better local optimum than when it is applied to networks with softmax. Therefore, when the objective function is highly non-convex, NCE can beat softmax even though it is only an approximation to softmax. % they  
%The statistical consistency of NCE and encouraging results shown above motivated us to performing a deeper analysis of NCE because in our view the potential of NCE has not been sufficiently explored for language modelling. 

Tuning hyperparameters has been an important element of neural networks research \cite{Bengio2012}. The main contribution of our paper is based on a carefully designed hyperparameter tuning strategy for NCE. The separate `search' and `convergence' phases for controlling the learning rate \cite{darken1991note} have never been applied to NCE-based neural networks. They appeared to be key components in this research, and they allowed NCE to outperform softmax on a problem on which softmax is known to have competitive performance and to be computationally feasible.

% In this work, we have done experiment on PTB dataset to see how we can apply NCE for language modelling efficiently. We have seen that choice of right hyper parameter is crucial for success of NCE. Although context dependent noise would make the performance better, it is difficult to use a local context dependent noise samples in GPU with mini-batch training \cite{Zoph2016SimpleFN}. From our experiment we have seen that zipfian(Zip) \cite{Piantadosi2014} distributions is also very expressive and can generate an efficient model.
%\todo[inline]{since we don't talk about contexts, we should probably comment this out. I have allready done that.}

% We have also proposed a new noise distribution which is based on global context of the word inspired by the Kneser-Ney smoothing technique.  %and does not hurt the computational efficiency with GPU. 

% We have achieved perplexity of {\bf69.975} on PTB benchmark with 7 hours training on Tesla k80 GPU which is comparable to the table~1 of \cite{zaremba2014recurrent}, where best result with single-model was {\bf78.4} with softmax.

Many researchers have probably concluded that NCE is not suitable for neural language modelling, and this was probably the reason why more sophisticated mechanisms, such as the `search' and `convergence' phases for controlling the learning rate, were not used with NCE. This seems to be a common pattern in deep learning research. For example, prior to 2016, the community believed that unsupervised learning has to be used to initialise supervised learning for neural networks, whereas today, the appropriate resources and engineering practices allow feedforward networks perform very well without unsupervised initialisation \cite[Ch.~6]{Goodfellow-et-al-2016}. Analogously, our paper shows that appropriate techniques exist to turn NCE into a successful method for language modelling.

The paper is organised as follows. Section~\ref{sec:back} introduces the NCE model with deep neural architecture, and Section~\ref{sec:model} describes our approach to NCE-based neural language modelling (NCENLM). Sections~\ref{sec:exp}~and~\ref{sec:res} describe the experimental design and the results showing that the proposed method improves the state-of-the-art results on the Penn Tree Bank dataset.

%% file: background.tex
%\todo[inline]{I think that explanation in the background should be improved. We can assume that people know more or less what backpropagation is and what a neural network is. Then we should provide an intuitive explanation what RNN is and then what LSTM is. At a high level. Assume that you are writing this for Alex. Then clearly define the input data tupes. I don't really understand $c_i$ right now. Note that you don't have a detailed plan for this section in the plan google doc. Perhaps you could first refine your plan in google doc, paragraph by paragraph, and then write this background section according to the detailed plan. Perhaps write backpropagation instead of BP.}
\begin{comment}
\begin{figure}[tb] %[!h]
\centering
\includegraphics[scale=0.75]{diagram_Lstm_NCE}
\caption{A three time step LSTM LM}
\label{diagram:lstm_lm}
\end{figure}
\end{comment}

We study language models where given a sequence of words $W= (w_1,w_2,\dots,w_T)$ over the vocabulary $V$, we model sequence probability
\begin{equation}
p(W) = \prod_{i=0}^{T-1} p(w_{i+1} | w_1, \dots, w_{i}) = \prod_{i=0}^{T-1} p(w_{i+1} |c_i).
\end{equation}
Here, for a given word $w_{i+1}$, $c_i = < w_1, \dots, w_{i}>$ represents its full, non-truncated context. %In practice, $w_0$ and $w_{T+1}$ are special symbols to denote the starting and ending of the sequence. 
%\info[inline]{following paragraph does not compile in latex??i commented out to compile}
In many applications, one is interested in $p(w_{i+1} |c_i)$. Recurrent neural networks try to model such probabilities that depend on a sequence of words $c_i$. The recurrent connections introduce a notion of `memory' which can remember a substantial part of word's context $c_i$. However, due to the gradient vanishing and exploding problems \cite{pmlr-v28-pascanu13}, it is challenging to optimise standard recurrent neural networks even though their expressive power is sufficient in many situations. For this reason, long short term memory (LSTM) was introduced to improve learning with a long context, $c_i$ \cite{hochreiter1997long,gers2001long}. LSTM introduces the concept of memory cells that are used to create layers. Several layers can be stacked into larger blocks (similar to layers of neurons in the multilayer perceptron). The blocks of those layers are then unrolled for several time steps during learning.
% a concept of a gating mechanism that can be seen as a complex unit.  n LSTM cell which, at a high level, can be seen as a complex neuron that can be used to create recurrent neural networks.
%Thus, LSTM cells can be stacked into a number of layers (similar to a multilayer perceptron) and they are usually unrolled for several time steps to improve learning of the historical contexts.
When $n$ is the last hidden layer, and $i$ is the last unrolled time step, $v^n_i$ is the activation vector that results after $c_i$ has been presented to the network. Then, the final output layer has one vector $\theta_j$ for every word $w_j$ in the vocabulary, and the probability of the next word can be computed using the softmax function:
%A N-layer recurrent neural network (RNN) with $n$ as the layer index and $i$ as the time step indicator in the unfolded RNN is modelled as $v_{i}^{n} = f_{RNN(\theta)}(v_{i-1}^{n}, v_{i}^{n-1})$, where $f_{RNN(\theta)}$ defines an affine transformation followed by nonlinearity, $v_i^{0}$ or $emb(w_i)$ represents the word $w_i$ at time step $i$ as 1-to-V encoding or dense vector encoding, $v_{i-1}^{n}$ denotes the recurrent vector representing the context until the time step $i$ and the $v_{i}^{n-1}$ represents the hidden vector from the $n-1$ layer of the time step $i$. Theoretically, a Recurrent Neural Network language model (RNNLM) can represent $c_i$ using the continuous vectors $v_{i-1}^{n}$ and $v_{i}^{n-1}$.  However, due to gradient vanishing and exploding problems \cite{pmlr-v28-pascanu13}, it is not clear exactly how much previous context is actually represented by the recurrent vector $v_{i-1}^{n}$.  For long and short term memory (LSTM) cell \cite{hochreiter1997long,gers2001long}, $f_{\theta}$ generates more sophisticated form of transformation than classic RNN using special gating mechanism which improves training by handling vanishing gradient problem better and thus captures longer dependencies. 
%Intuitively, LSTM can store and retrieve information over long time scales using its gating mechanisms. 
%The Unnormalised output from the top layer of the last unfolded LSTM cell, which does not provide any probability distribution, will be feed into a normalisation process (through softmax) as follows:
\begin{equation}
P_{\theta}^{SOFT}(w_{i+1} | c_i)=\frac{\exp(\theta^{\top}_{i+1} v_{i}^n)}{\sum_{j=1}^{|V|}\exp(\theta^{\top}_{j} v_{i}^n)}=\frac{\exp(\theta^{\top}_{i+1} v_{i}^n)}{Z}.
\label{eq:softmax}
\end{equation}
Here, $P_{\theta}^{SOFT}(w_{i+1} | c_i)$ is the probability of word $w_{i+1}$ given context $c_i$, $\theta_{i+1}$ is the weight vector corresponding to the word $w_{i+1}$ at the output layer, $\theta_{j}$ is the weight vector for the word $w_j$ in vocabulary%, $v_{i}^n$ is the recurrent vector representing the context propagated through the hidden layer of $n^{th}$ layer
, and $|V|$ is the vocabulary size. The normalising term $Z$ is known as the {\bf partition function}. Note that unnormalised products $\theta^{\top}_{i+1} v_{i}^n$ are not sufficient to evaluate the words.

%Lets define the partition function $Z = \sum_{j=1}^{|V|}\exp(\theta^{\top}_{j} v_{i-1})$, for normalising the the probability distribution $ P_{RNN}$. 
%unnormalized score
%$$v_j = \tanh(x \times W^1) \times W_{*,j}^2$$
%for $j \in \{1, \dots, |v|\}$
% We use softmax to force a distribution. 
%$$Softmax (z_j) = \frac{\exp(z_j)}{\sum_{w \in v}{\exp(z_w)}}$$

The softmax-based training of recurrent neural networks that uses stochastic gradient descent (SGD) and backpropagation (BP) maximises the log likelihood or equivalently minimises the cross-entropy of the training sequence containing $N$ words. This objective can be formally expressed as 
\begin{equation}
J_{SOFT}(\theta) = - \frac{1}{N} \sum_{i = 1}^{N} \ln P_{\theta}^{SOFT}(w_{i+1} | c_i).
\label{eq:ce_loss_softmax}
\end{equation}
The gradient used for updating the parameters $\theta$ is
\begin{equation} \label{eq:soft_ce_loss_grad}
\begin{aligned}
\frac{\partial J_{SOFT}(\theta)}{\partial \theta} & =  - \frac{1}{N} \sum_{i = 1}^{N} \Big[ \frac{\partial (\theta_{i+1}^{\top} v_{i}^n)}{\partial \theta}  \\  & - \sum_{j=1}^{|V|} P_{\theta}^{SOFT}(w_j | c_i) \frac{\partial (\theta_{j}^{\top} v_{i}^n)}{\partial \theta} \Big].
\end{aligned}
\end{equation}
Gradient computation is usually time-consuming because when the vocabulary is large, the partition function in $P_{\theta}^{SOFT}$ creates the performance bottleneck for the training and testing phases. It is advantageous to avoid this expensive normalisation term.
%\info{The rare words does not appears all the time, the gradient contribution should be handled differently from the rare words, in mini-batch how the combination of rare and frequent words make a combined contribution}. 
Noise contrastive estimation (NCE) bypasses this calculation by converting the original optimisation problem to a binary classification problem.

In NCE, we see the corpus as a new dataset of $n$ words of the following format: %n = sequence size
\begin{equation*}
((c_1,w_2), D_1)), \dots , ((c_n,w_{n+1}),D_n)
\end{equation*} 
where $c_i$ represents the context, %$w_{i-n+1}, \dots, w_{i}$
$w_{i+1}$ represents the next word after $c_i$, and a random variable $D$ is set to one when $w_{i+1}$ is from the training corpus (true data distribution) and $D$ is set to zero when $w_{i+1}$ is from a known chosen noise distribution, $P_n$. For a given context $c_i$, the NCE-based neural language model (NCENLM) models data samples (from the corpus) as if they were generated from a mixture of two distributions ($P_{\theta}^{NCE}$ %~\footnote{assuming $P_{\theta}^{NCE}$ can approximate the true empirical distribution} 
and $P_n$). The mixture is normalised; hence, the requirement for the normalisation term is satisfied implicitly as shown in Eq.~(\ref{eq:posterior_p_data_p_n}). %In NCE, for each data item $w_{i+1}$, $k$ noise samples are generated, and \cite{Mnih12afast} shows that for large $k$, NCE-based parameter estimation is a close approximation of the MLE.
 
The posterior probability of a sample word $w_{i+1}$ generated from the mixture of the $P_{\theta}^{NCE}$ and the noise distribution $P_n$ are as follows:
\begin{equation} \label{eq:posterior_p_data_p_n}
\begin{aligned}
P(D = 1| w_{i+1} , c_i) = \frac{P_{\theta}^{NCE}(w_{i+1}| c_i)}{P_{\theta}^{NCE}(w_{i+1}|c_i) + k P_n(w_{i+1}|c_i)} \\
P(D = 0| \widetilde{w_{i+1}} , c_i) = \frac{k P_n(\widetilde{w_{i+1}}|c_i)}{P_{\theta}^{NCE}(\widetilde{w_{i+1}}|c_i) + k P_n(\widetilde{w_{i+1}}|c_i)}
\end{aligned}
\end{equation}
where $\widetilde{w_{i+1}}$ is a word sampled from a known noise distribution $P_n$, e.g., a uniform distribution. Based on this posterior distribution, NCE minimises the following objective function:
\begin{equation} \label{eq:nce_loss}
\begin{aligned}
J_{NCE}(\theta) &= - \frac{1}{N} \sum_{i=1}^{N} \Big[ \ln P(D = 1| w_{i+1} , c_i) \\
& + \sum_{j=1}^{k} \ln P(D = 0| \widetilde{w_{i+1,j}} , c_i) \Big].
\end{aligned}
\end{equation}
which is the same objective function (up to a factor of $\frac{1}{2}$) that is minimised by the traditional logistic regression. Here, for every word $w_{i+1}$ that comes from a true data distribution, $k$ noise samples  $\widetilde{w_{i+1,j}}$ are generated from a known noise distribution, $P_n$. \cite{Mnih12afast} show that for large $k$ NCE-based parameter estimation is a close approximation of the maximum likelihood estimation.
% \info{check last lines of page 298 of the Michael Gutmann's 2010 paper; Noise-constrastive estimation: a new estimation principle for unnormalised statistical models.}

The gradient of the objective function is as follows:
\begin{equation} \label{eq:nce_loss_grad}
\begin{tiny}
\begin{aligned}
& \frac{\partial J_{NCE}(\theta)}{\partial \theta} = \\
& - \frac{1}{N} \sum_{i=1}^{N} \Bigg[ \frac{k P_n (w_{i+1}|c_i)}{P_{\theta}^{NCE} (w_{i+1}|c_i)+k P_n(w_{i+1}|c_i)} \frac{\partial}{\partial \theta} \ln P_{\theta}^{NCE} (w_{i+1}|c_i) \\
& - \sum_{j=1}^{k} \frac{P_{\theta}^{NCE}(\widetilde{w_{i+1,j}}|c_i)}{P_{\theta}^{NCE}(\widetilde{w_{i+1,j}}|c_i) + k P_n(\widetilde{w_{i+1,j}}|c_i)} \frac{\partial}{\partial \theta} \ln  P_{\theta}^{NCE}(\widetilde{w_{i+1,j}}|c_i) \Bigg] 
\end{aligned}
\end{tiny}
\end{equation}
where, 
\begin{equation}\label{eq:P_rnn_in_nce}
\begin{aligned}
P_{\theta}^{NCE}(w_{i+1} | c_i) = \frac{\exp(\theta^{\top}_{i+1} v_{i}^n)}{Z} \\
P_{\theta}^{NCE}(\widetilde{w_{i+1,j}} | c_i) = \frac{\exp(\theta^{\top}_{j} v_{i}^n)}{Z}.
\end{aligned}
\end{equation}
In the softmax gradient in Eq.~\ref{eq:soft_ce_loss_grad}, the normalisation therm $Z$ is required to compute $P^{SOFT}_\theta$, which is a problem during training because $Z$ has to be computed for every gradient calculation. Studying the NCE gradient in Eq.~\ref{eq:nce_loss_grad}, one can see a subtraction of two large products. The first terms of those products are normalised implicitly regardless $P^{NCE}_\theta$ is normalised or not. Thus, the only terms in which normalisation can matter are the gradients in Eq.~\ref{eq:nce_loss_grad}. It has been argued in the literature, however, that as far as the gradient is concerned, $Z$ can be learnt as a parameter \cite{gutmann2010noise} or it can be seen as a constant, for instance, $Z=1$ was used for all contexts in \cite{Mnih12afast,conf/emnlp/VaswaniZFC13,Zoph2016SimpleFN}. This is the precise reason why the partition function $Z$ does not have to be computed in every iteration in NCE.

%% file: nce_training.tex
In this section, we present our training procedure with special hyperparameter tuning for NCE-based neural language models (NCENLM). For training the model, stochastic gradient descent (SGD) is used because SGD yields significantly better generalisation than batch methods \cite{bousquet2008tradeoffs}, specially when an appropriate learning rate schedule is utilised. The following hyperparameters turned out to be important for our NCE-based language model: the learning rate schedule, the dropout rate, and the weight initialisation strategy. In our paper, the words are represented as the word vectors  \cite{NIPS2013_5021} trained using the standard continuous skip-gram model with negative sampling. %\todo{I don't understand this sentence. Is the negative sampling algorithm part of the representation or is the algorithm used to compute the representation? Is the skip-gram used instead of word2vec vectors? How does it work? Ans: Word's vector representation is learned  using the skip-gram with negative sampling algrithm. So it is the algorithm used to compute the representation. There is another variation skip-gram with hierarchical softmax. Skip-gram did not used instead of word2vec. However, when people say they have initialized with word2vec, they means that words were learned using one of this algorithms} 
%The dropout probability also played important rule in regularising the network with small dataset.

% \todo{Can we use parameters instead of hyperparameters? Why do we have "hyper-"?Ans: Parameters are the weights of the neural model, hyperparameters are variables set before actually optimizing the model's parameters (weight), like learning rate is a hyperparameter, we have to set learning rate before we can start training the network with SGD to get optimal weights of neuron's connection(parameters).}

%Utilising the efficient embeddings with NCENLM gave us performance enhancement(better PPL) compared to initialising the first hidden layer with random scaled value and learning the embedding from the NCENLM training (see Table~\ref{tab:nce_no_vec}).  

%\subsection{Word Representation} \label{sec:word-rep}
%In this paper, the words are represented with dense vector representation using skip-gram with negative sampling \cite{mikolov2013efficient} algorithm. 

\subsection{Learning rate} \label{sec:learning-rate}

%figure was here

\input{learning_rate}

\subsection{Weight Initialisation} \label{sec:init-weights}
%{\bf Impact of weight initialisation with Sigmoid activation:}

%\input{weight_initialization_sigmoid}

%{\bf Weight Initialisation for NCE}
%\info{Formulating for NCE}

%\improvement{What happened with tanh activation?Observe the images from the different initialisation. See their variance and mean.}
\input{weight_initialization_nce}

%\subsection{Regularisation} \label{sec:dropout}
%\input{reguraliser}

%% file: learning_rate.tex
The learning rate is one of the most prominent hyperparameters in deep network training \cite{Bengio2012}. The `search-then-converge' learning rate schedule for SGD usually has the form of $\eta(t) = \eta_0 (1 + \frac{t}{\tau})^{-1}$ \cite{darken1991note}, where $t$ is the epoch number, $\eta_0$ is the initial learning rate, $\tau$ is a parameter, and $\eta(t)$ is the learning rate for epoch $t$. This allows the learning rate to stay high during the `search period' $t\leq\tau$. It is expected that during this period the parameters will hover around a good minimum. Then, for $t>\tau$, the learning rate decreases as $\frac{1}{t}$, and the parameters converge to a local optimum because this schedule agrees with the stochastic approximation theory \cite{robbins1951stochastic}. %This power law schedule achieves the optimal $\frac{1}{t}$ convergence rate. From the stochastic approximation theory \cite{robbins1951stochastic}, we find the conditions on the learning rate to ensure optimal fast convergence. The suggested learning rate sequence has the form $\eta(t) = \eta_0(\frac{1}{t})$ for $\eta_0 > 0$.

In our implementation, we used the `search-then-converge' learning rate schedule of the form:
\begin{equation}\label{EQ:LRSCHEDULE}
\eta(t) = \eta_0 \times \Big({\frac{1}{\psi}}\Big)^{\max(t+1-\tau, 0.0)},
\end{equation}
which previously appeared in other studies that involve neural networks \cite{zaremba2014recurrent}. The hyperparameter $\psi$ is kept constant in our experiments and its value was set according to \cite{zaremba2014recurrent}. During the search period ($t\leq\tau$ epochs), the learning rate is constant and equal to $\eta_0$, and during the convergence period the learning rate is decreased by a factor of  $\frac{1}{\psi}$. The initial learning rate $\eta_0$ is one in \cite{zaremba2014recurrent}, and we use the same value in our experiments. %\todo{please check the sentence above; what is the value of $\psi$ in our experiments? Ans:We have used the same value as used in the zaremba2014recurrent paper. For small, medium and large, the values are 0.5,0.8,0.87. This is another hyperparameter and it decides the learning rate decay. I have found that for softmax and NCE the same value worked fine.}

\textit{Our investigation has shown that learning with NCE is more sensitive to the length of the search period ($t\leq\tau$) when comparing with softmax. Choosing appropriate $\tau$ is, therefore, crucial for convergence of NCE-based learning with SGD. Our research suggests that, when NCE is used, $\tau$ should be between 1 and two-thirds of the total number of training epochs. For instance, if we need 40 training epochs then $\tau$ could be between 1 and 26. This was one of the most important insights that allowed us to improve the performance of NCE.}

%% file: weight_initialization_nce.tex
In neural networks, the initial weights are usually drawn from a uniform distribution \cite{glorot2010understanding}, unit Gaussian \cite{sutskever2013importance} or a general Gaussian distribution \cite{he2015delving}. %These papers suggest values of variance for the distributions from which the initial weights are drawn. The variance depends on the activation function used and it is derived without explicitly considering the type of the distribution. As such, these recommendations hold for any type of distribution.  
For our NCE training, we used a uniform distribution for the weight initialisation. All three distributions described above were compared, but the uniform distribution led to slightly better results. \textit{However, regardless which distribution is used, we found that NCE works much better when the initial weights are within a smaller range, i.e. when the variance of the initial weights is smaller than what is suggested in \cite{glorot2010understanding}. This was another insight that led to significant improvements in NCE performance.}

%If we study the initial training perplexities, then we see that initially the model is extremely perplexed (very high PPL value). %Although the output layer is less influential for weight initialisation, at the backward pass, the gradient has impact on the weight change. 
%In NCE, the entire weight matrix of the output layer (hidden to output) does not change during the backward pass. Only the part of the weight matrix corresponding to the sampled noise data gets updated in each iteration. As the back-propagated gradients were smaller as one moves from the output layer towards the input layer, just after initialisation \cite{bradley2010learning}, possibly initialising weights with smaller variance helps keeping the consistency in weight distribution.

%% file: table_compare_state_of_the_art.tex
\begin{table*}[tb]
\begin{center}
\caption{Comparison with the state-of-the-art results of different models on the PTB dataset}
\label{table:compare_state_of_the_art}
{\small
 \begin{tabular}{|c || c | c|} 
\toprule
  \multicolumn{3}{c}{Classic RNN and LSTM} \\
\bottomrule \hline
   Model Description & Validation PPL & Test PPL \\ [0.5ex] 
 \hline
 Deep RNN  \cite{DBLP:journals/corr/PascanuGCB13} &-& 107.5\\
 \hline
Sum-Prod Net  \cite{cheng2014language}&-& 100.0 \\
  \hline
  RNN-LDA + KN-5 + cache  \cite{mikolov2012context} &-& 92.0\\
 \hline
 Conv.+Highway+ regularized LSTM \cite{kim2016character} & - & 78.9 \\

\hline\hline
 Non regularised LSTM with Softmax \cite{zaremba2014recurrent}& 120.7 & 114.5 \\  
 \hline
 Medium regularised LSTM with Softmax \cite{zaremba2014recurrent} & 86.2  & 82.7 \\
 \hline
 Large regularised LSTM with Softmax \cite{zaremba2014recurrent}& 82.2 & 78.4\\
 \hline\hline
 Non regularised LSTM with NCE (our method) & 106.196 &102.245\\  
 \hline
 Medium regularised LSTM with NCE (our method) &  78.762 & 75.286\\
 %fine tuning
 \hline
 Large regularised LSTM with NCE (our method) & 72.726 &  {\bf 69.995} \\
 %rnn/Zip_Ascending_order/result_correct/L_Abs_ascending_ppl_69.975.txt
 \hline

 \toprule
  \multicolumn{3}{c}{Extended or Improved LSTM} \\
\bottomrule \hline
 Variational LSTM \cite{gal2016theoretically} &77.3 & 75.0\\
\hline
Variational LSTM + WT  \cite{DBLP:journals/corr/PressW16} & 75.8 & 73.2 \\
\hline
Pointer Sentinel LSTM \cite{DBLP:journals/corr/MerityXBS16} &72.4 & 70.9\\
\hline
Variational LSTM + WT + augmented loss \cite{DBLP:journals/corr/InanKS16} & 71.1 & 68.5\\
\hline
Variational RHN \cite{DBLP:journals/corr/ZillySKS16} & 71.2 & 68.5 \\
\hline
Variational RHN + WT \cite{DBLP:journals/corr/ZillySKS16}  & 67.9 & 65.4 \\

\hline
Neural Architecture Search with base 8 and shared embeddings &-&62.4\\
Utilises a novel recurrent cell and reinforcement learning \cite{zoph2016neural} &&\\
\hline
\toprule
  \multicolumn{3}{c}{Model Averaging/ Ensembles} \\
\bottomrule \hline

38 large regularized LSTMs \cite{zaremba2014recurrent} &71.9 &68.7\\
\hline
Model averaging with dynamic RNNs and n-gram models  \cite{mikolov2012context}&-&
72.9 \\
\hline
\end{tabular}
}
\end{center}

%\label{table:compare_state_of_the_art}
\end{table*}

%% file: experiment.tex
We aim at showing that NCE can outperform alternative methods for language modelling. In particular, we investigate its performance in the context of softmax because NCE approximates softmax being consistent with softmax in the limit. Our implementation of NCE follows our approach presented in Sec.~\ref{sec:model}.

In our experiments, we focus on the popular perplexity measure (PPL) using the Penn Tree Bank (PTB) dataset. It is feasible to run `exact' softmax on this dataset, and the large literature that uses it allows for comparisons with other approaches (see Tab.~\ref{table:compare_state_of_the_art}).
% We evaluated the performance of the NCENLM described in Sec.~\ref{sec:back} utilising the training methodology described in Sec.~\ref{sec:model}, using the popular perplexity measure on Penn Tree Bank (PTB) dataset. 
% \footnote{http://www.fit.vutbr.cz/~imikolov/rnnlm/simple-examples.tgz} \cite{marcus1993building}. 
The PTB dataset consists of 929k training words, 73k validation words, and 82k test words. The vocabulary size is 10k. Softmax usually becomes inefficient when the vocabulary size exceeds 10k words. % Language modelling with a vocabulary size of more than 10k words is known as large scale language modelling where softmax becomes inefficient. % We have used PTB for this experiment where softmax is known to be efficient. 

%We wanted to show that NCE can do better than Softmax where softmax is strong in performance. 

% \cite{abadi2016tensorflow}
All models were implemented in Tensorflow\footnote{\url{https://www.tensorflow.org/}} and executed on NVIDIA K80 GPUs. The standard components of our models follow \cite{zaremba2014recurrent} where excellent results on this dataset were reported. %We have used most of the hyperparameters values from \cite{zaremba2014recurrent} with some exception in the learning rate schedule, dropout rate, and weight initialisation.
The words were represented with dense vectors trained using the skip-gram model with negative sampling \cite{NIPS2013_5021} on the Wikipedia's corpus downloaded on December 2016. This word representation was used across all our experiments. % \todo{If I were a reviewer, I would argue that the great perplexity that was achieved was due to better word vectors? Did the other people whose results are in table 1 use the same dataset to train word vectors?Ans: But the softmax was trained using the same word2vec in table2. Does it answer the reviewers question?}
%We also did experiment with skip-gram with hierarchical softmax and got similar results.
 
% During the experiment, we have fine tuned the word vectors for medium and large model to observe the fine-tuning impact on the  language model.
% \unsure{the above paragraph can be removed? I removed this paragraph. We would need more space to explain this stuff.}

In order to perform more experiments, we designed models of three sizes: small (S), medium (M) and large (L). The small model is non-regularised whereas the medium and large models are dropout regularised with 50\% and 60\% dropout rate on the non-recurrent connections in the medium and larger models correspondingly. This led to the best empirical results after investigating different dropout rates in the suggested ranges \cite{srivastava2014dropout}. All the models have two LSTM layers with the hidden layer size of 200 (S), 650 (M), and 1500 (L). The LSTM was unrolled for 20 time steps for the small model and 35 time steps for the medium and large models. We used mini-batch SGD for training where the mini batch size was 20.

\begin{figure}[h]
\centering
\includegraphics[width=0.4\textwidth,scale=0.5]{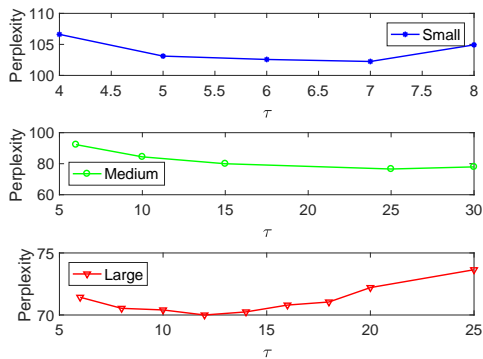}
\caption{Selection of the learning rate parameter $\tau$}
\label{fig:learning_rate_annealing}
\end{figure}

For sampling the initial weights, a smaller range than the one suggested in \cite{glorot2010understanding} turned out to be very efficient for NCE. We tested several initialisation heuristics which are described in the corresponding column in Tab.~\ref{tab:weight_init_nce}. Row number 1 shows the formula suggested in \cite{glorot2010understanding}. Note that $U$ denotes a uniform distribution with its minimum and maximum values. %and corresponding test PPL for three models. 
Row number 2 shows our updated formula that reduces the first range by a factor of 4, and row number 3 shows the range values that led to the best results in our experiments. %We have drawn weights from the uniform distribution with a higher variance than the standard used in literature. 
%\todo{what does this mean U(-0.0153,0.0153)? Is it variance in a Gaussian distr. or min max in a uniform distrib.? Ans: It is from uniform distribution, specifically random uniform initializer. The variance for $U(a,b) = (b-a)^2/12$. We are not specifying the variance explicitly here. In Tensorflow, if we use the initialiser from a specific range, for random uniform, U(a,b), then the weight will be initialize from random uniform distribution from this range} \todo{if the rows were numbered in tab. 3, it would be easier for people to read the descr. Ans: OK, Done}

The learning rate was scheduled using Eq.~\ref{EQ:LRSCHEDULE}. The search time limit $\tau$  was chosen empirically using Fig.~\ref{fig:learning_rate_annealing}.
%We train the LSTM for 39 epochs with learning rate 1, after 25 epochs we decrease the learning rate by a factor of 1.25 \unsure{check for correctness}. 
As a result, $\tau$ was set to 7, 25 and 12 for the small, medium and larger models correspondingly. During the convergence period, the parameter $\psi$ was set to 2, 1.2 and 1.15 for the small, medium and larger models as suggested by \cite{zaremba2014recurrent}. We trained the models for 20, 39 and 55 epochs respectively.

%Although we found that M and L model can be trained with 39 epochs without loss of efficiency in terms of perplexity.  
%\info{Explain this fast learning in terms cross entropy, larger error make learning faster}
The norm of the gradients (which was normalised by the mini batch size) was clipped at 5 and 10 for the medium and large models correspondingly. 
%The training+testing took 2 hours and 6 hours respectively on NVIDIA K80 GPU. For small non-regularized model, we have used training with learning rate 1 for first 7 epochs (search time) and  learning rate decay $\psi$ of 0.5 was used in the converge time. We train the model for 20 epochs. 
To compute validation and testing perplexity, we used softmax to guarantee accuracy of our comparisons. %It took 30 mins to train+test on the NVIDIA K80 GPU. For testing we used softmax to make comparison fair with other methods in perplexity wise.\info{find the paper where they used NCE in testing} 

%We have used different noise sampling as described in section ~\ref{sec:sampling_tech}. 
In NCE, we used 600 noise samples. The noise samples were generated from the power law distribution. We evaluated different noise sample sizes (50, 100, 150, 300, 600, and 1200), and 600 had the best trade-off between quality and time. We observed that when a GPU implementation is used, it is possible to increase the sample size within a reasonable range without dramatically increasing  the computational complexity. % We observed that the noise sample size does not increase the training time extensively. It depends on the GPU capacity. If we have access to powerful GPU, we can increase the sample size. 

%\unsure{add a graph showing the number of noise sample with ppl?} 

%\info{as much as they can fit in a GPU's cache :8192, if the vocabulary size is more than this 8192 then softmax become really expensive}

% For comparing the efficiency with softmax, we have also run the experiments training+validation+testing with the softmax output layer, where we used parameterisation
Our softmax-based language model was implemented and parametrised accroding to \cite{zaremba2014recurrent} where it achieved the state-of-the-art results using the classic LSTM cell.
%\info{controlled experiment}

The two models that we implemented, i.e. softmax- and NCE-based language models, used a classic LSTM cell \cite{hochreiter1997long,gers2001long}. %which gave state-of-the-art results in LM. 
%knowing that LSTM cannot solve entirely the long term dependency problem for complex problem like LM.
Many extensions to the LSTM architecture exist, e.g., Recurrent Highway Networks (RHN) \cite{DBLP:journals/corr/ZillySKS16}, that may improve LSTM's capabilities in capturing long term dependencies. %solve this long term dependency problem which would improve the language modelling performance as well. #
In this paper, we aimed at comparing NCE and softmax on standard LSTM networks, but our results could generalise to other, potentially more advanced types of LSTM cells. It should be noted, however, that our NCE implementation with standard LSTM outperforms some language models which use more advanced versions of LSTM as shown in Tab.~\ref{table:compare_state_of_the_art}. %We have worked with classic LSTM cell \cite{hochreiter1997long,gers2001long} and empirically observed how the normalisation of the output layer has impact on language modelling performance. %More specifically, we will compare between normalised and unnormalised model.  
%\todo[inline]{It seems that the second part of the paragraph above could be moved to the end of section 4. In section 4 you describe methodology. In there you could say that you use state-of-the-art LSTM, but you acknowledge that improvements exist and then you mentioned a few of them. You could justify the type of LSTM that you are using, an by adding these citations you are showing that you are aware of the existing extensions.}
%\info[inline]{OK?}
%For the Zipfian distribution, the words in the vocabulary have to be sorted in descending order. In general, for many application, this can't be possible. We have done experiments where the words in vocabulary was sorted in ascending order and where the vocabulary is not sorted in any order.  

%Our intension is to show that we can achieve the state-of-the-art result using non zipfian distribution.

%% file: results.tex
%\info{Does NCE take less parameter? hyperparamters are a confounding factor at https://drive.google.com/file/d/0B2A1tnmq5zQdRDQ1d29jNk5MOUk/view}
%%%%%%%%%%%%%%%%%%%%

\begin{table}[tb]
\begin{center}
\caption{Comparison of softmax and NCE}
\label{table:nce_time}
{\small
 \begin{tabular}{|c || c | c|c |} 
%\toprule
%  \multicolumn{3}{c}{} \\
%\bottomrule
 \hline
   Large Model & Time & Valid. PPL & Test PPL\\ [0.5ex] \hline\hline
   Softmax (55 epochs) & 9 h 11 min & 82.588 & 78.826\\\hline
   Softmax (20 epochs) & 3 h 40 min & 79.798 &  76.935\\\hline
   NCE (55 epochs)& 7 h 34 min &  72.726  &  69.995 \\\hline
   %NCE (ep no=40) & 5 hr 46 min & 73.203& 70.240\\\hline
   NCE (20 epochs)& 2 h 36 min &76.268    &  74.129  \\\hline
\end{tabular}
}
\end{center}
\end{table}

%%%%%%%%%%%%%%%%%%%%%%%%%%%%%%%%

%%%%%%%%%%%%%%%%%%%%%%%%%%%%%%%%%%

\begin{table*}[tb]
\begin{center}
\caption{Weight initialisation ranges for the uniform distribution (U) and the corresponding test perplexity (PPL)}
\label{tab:weight_init_nce}
{\small
 \begin{tabular}{|c||c || c | c|c|} 
%\toprule
%  \multicolumn{3}{c}{} \\
%\bottomrule 
\hline
   No.&Initialisation Heuristic & Small Model & Medium Model & Large Model \\ [0.5ex] 
 \hline
 1&$U \left( - \frac{\sqrt{6}}{\sqrt{n_{i} + n_{i+1}}} , \frac{\sqrt{6}}{\sqrt{n_{i} + n_{i+1}}}\right )$ &U(-0.1225, 0.1225)&U(-0.0679, 0.0679)& U(-0.04472, 0.04472)\\
 
& &PPL = 104.449&PPL = 75.960&PPL = 71.184\\
 \hline
  2&$ U \left( - \frac{\frac{\sqrt{6}}{\sqrt{n_{i} + n_{i+1}}}}{4} , \frac{\frac{\sqrt{6}}{\sqrt{n_{i} + n_{i+1}}}}{4} \right )$ &U(-0.031, 0.031)& U(-0.0169, 0.0169) &U(-0.011180, 0.011180) \\
  &&PPL = 102.245&PPL = 75.959&PPL = 70.444\\
  \hline
 %Arbitrary& U(-0.1,0.1)&$U(-0.05, 0.05)  $  & U(-0.00625, 0.00625)\\
 %&105.796&75.966& 70.244(z) 69.975(u(a)) \\
 
  3&Empirically Tuned Ranges & U(-0.0153, 0.0153)&$U(-0.00849, 0.00849)  $  & U(-0.00625, 0.00625)\\
 && PPL = 102.237 & PPL = 75.286 & PPL = 69.995 \\
 \hline
\end{tabular}
}
\end{center}
\end{table*}

%% file: result_discussion.tex
\begin{figure}[tb]
\centering
\includegraphics[width=0.4\textwidth,scale=0.5]{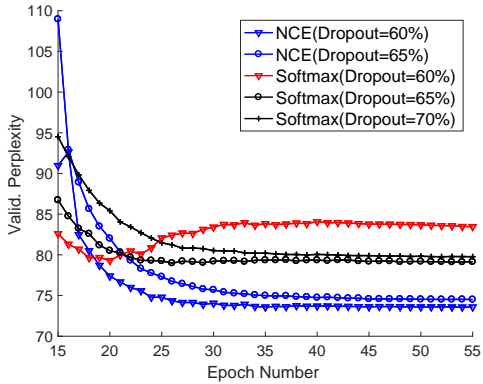}
\caption{Convergence phase in the large model}
\label{fig:convergence_large}
\end{figure}

The results in Tab.~\ref{table:compare_state_of_the_art} compare our best NCE-based result with other state-of-the-art methods. Our result is the best in the class of single-model methods; the large model achieved the perplexity of {\bf 69.995} after 55 training epochs. This result outperforms all known single-model algorithms that use the same kinds of LSTM cells. % For completeness, we list several results on modified RNN/LSTM and ensembles. % With a classic LSTM cell, unnormalised modelling with NCE can achieve the highest score (lowest perplexity) if the hyper parameters are chosen correctly. 
%With the L model, NCENLM achieved the perplexity of {\bf 69.995} which is better than any single-model could do until now. %where noise samples were drawn from Zipfian distribution. 
The total time for training, validating and testing our large NCE-based model was 7 hours 34 minutes (see Tab.~\ref{table:nce_time}). %\todo{does testing use softmax to compute perplexity? so maybe we lose a lot of time on softmax in there? is it how people compare these algorithms? Ans: Yes. validation and testing was in softmax. But we want to compare against the softmax, so we have to keep then in softmax, Right?} %We also run the experiment with less number epochs of 39 and got perplexity of 70.240 which took 5 hr 46 min. 
The 55 epochs of softmax took 9 hours 11 minutes, and the testing perplexity was 78.826. Early stopping, which is a common regularisation method \cite{Goodfellow-et-al-2016}, allowed softmax to achieve a testing perplexity of 76.935. So, softmax was clearly overfitted after 55 epochs. The same overfitting was not observed in NCE as can be seen in Tab.~\ref{table:nce_time} and Fig.~\ref{fig:convergence_large}. % As the large model become overfitted on this dataset, we used the weights from the best validation perplexity and the corresponding test perplexity was 76.935. %\todo{the last sentence is fishy; does it mean that we did not put enough effort to tune the best softmax result? could we do something here to avoid overfitting in softmax?Ans: we tried to use dropout to avoid overfitting, see figure 2 soft\_L\_70 plot where softmax did not got overfitted but generalization reduces and test PPL was worse}

Below, we present additional results that explain good performance achieved by NCE and provide further insights into its properties.

%large model analysis with dropout
Figure~\ref{fig:convergence_large} presents the validation perplexity (Y axis) of a large model for different dropout rates as a function of an epoch number (X axis) at the convergence stage of learning. One can see that softmax with a dropout rate of 60\% overfitted since the 21st epoch. Increasing the dropout rate to 70\% allowed softmax to avoid overfitting, but the asymptotic performance was not as good as in NCE. The asymptotic convergence of NCE was superior across a range of dropout rates. %the model gave better results with the same dropout rate and did not overfit.
In NCE, the gradients (Eq.~\ref{eq:nce_loss_grad}) are different and more noisy than in softmax (Eq.~\ref{eq:soft_ce_loss_grad}). We know that SGD leads to better generalisation than batch gradient descent because of the induced noise by updating the parameters from a single example \cite{bousquet2008tradeoffs}. Similar property of NCE could justify its robust generalisation in Fig.~\ref{fig:convergence_large}.
%\todo{What does L and LT in Figure 2 mean? What is the different between lines that have L from those that have LT?Ans: Ok, LT means when the embeddings were fine tuned, we can remove this if we dont discuss fine tuning in section 4 and 5. I keep these for the time being but if we need space reduction then I will remove them.}

%medium
\begin{figure}[t]
\centering
\includegraphics[width=0.4\textwidth,scale=0.5]{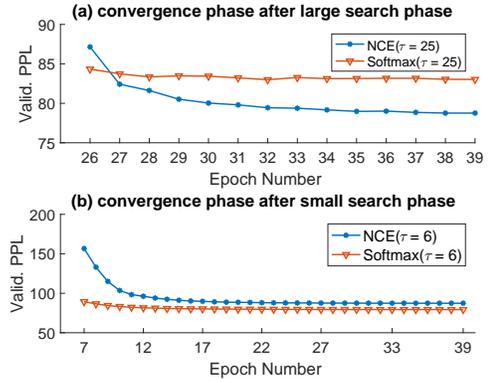}
\caption{Convergence phase in the medium model}
\label{fig:convergence_phase}
\end{figure}
%large
\begin{figure}[t]
\centering
\includegraphics[width=0.40\textwidth,scale=0.5]{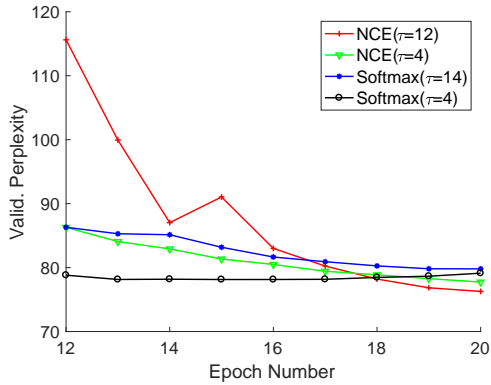}
\caption{Convergence phase in the large model}
\label{fig:convergence_phase_L}
\end{figure}

\begin{comment}
\begin{figure*}[tb]
\centering
\begin{subfigure}{.5\textwidth}
\centering
\includegraphics[width=.5\linewidth]{convergence_phase}
\caption{Convergence Phase for M model with different $\tau$}
\label{fig:convergence_phase}
\end{subfigure}%
\begin{subfigure}{.5\textwidth}
 \centering
\includegraphics[width=.5\linewidth]{large_convergence_12_20}
\caption{Convergence Phase(L)}
\label{fig:convergence_phase_L}
\end{subfigure}
\caption{A figure with two subfigures}
\label{fig:test}
\end{figure*}
\end{comment}

\begin{figure}[h]
\centering
\includegraphics[width=0.4\textwidth,scale=0.5]{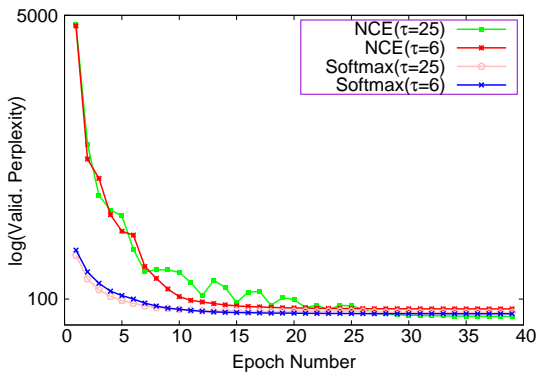}
\caption{Validation perplexity of the medium model during all epochs of learning}
\label{fig:early_high_error}
\end{figure}

Figures~\ref{fig:convergence_phase}~and~\ref{fig:convergence_phase_L} show the validation perplexity (Y axis) for selected values of $\tau$ as a function of an epoch number (X axis) at the convergence stage of learning. These figures demonstrate the critical impact of the learning rate schedule on NCE. Figure~\ref{fig:convergence_phase} shows that NCE requires a long search period (large $\tau=25$) to achieve competitive asymptotic convergence on the medium model. Figure~\ref{fig:convergence_phase_L} for the large model has additional evidence that a long search period is required because larger $\tau=12$, in addition to having better asymptotic convergence, has poor (i.e. high) perplexity in the initial phase. This poor perplexity indicates that the algorithm explores widely at this stage, but by doing that it can avoid converging to the nearest local optima. High initial perplexity is even more pronounced in Fig.~\ref{fig:early_high_error} which is for all epochs of the medium model (note that perplexity is on the log scale here). Although difficult to see in the figure, the asymptotic validation perplexity is the best for NCE with $\tau=25$. There was also a difference in test performance between NCE and softmax: NCE with $\tau=25$ scored 75.959, NCE with $\tau=6$ scored 83.858, softmax with $\tau=25$ scored 79.906, and softmax with $\tau=6$ achieved 78.567. NCE with high $\tau$ was clearly the best, and increasing $\tau$ from 6 to 25 reduced perplexity from 83.858 to 75.959, which confirms the significance of our arguments in Sec.~\ref{sec:model}. %From these two figures, we can see that convergence is faster for NCE after large search phase ($\tau = 25$ in (a) compared to $\tau = 6$ in (b) at Fig.~\ref{fig:convergence_phase} and $\tau = 12$ compared to $\tau = 4$ in Fig.~\ref{fig:convergence_phase_L}) was employed.
%Again, the relationship between stochastic and batch gradient descent can explain this behaviour of NCE. 
Thanks to the noise samples, NCE can explore better than softmax when the exploration phase is long enough which is confirmed through high perplexity in the initial stage of learning. This means that NCE can find a better solution potentially for the same reasons which make stochastic gradient descent better than batch gradient descent \cite{bottou2010large}.

\begin{figure}[h]
\centering
\includegraphics[scale=0.6]{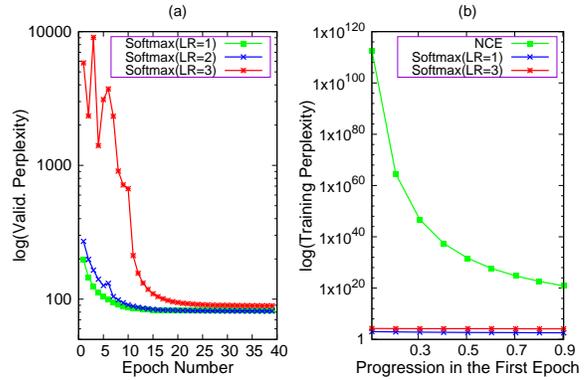}
\caption{High learning rate (LR) to increase the initial softmax perplexity (a). NCE and softmax initial perplexities in the first epoch; only training perplexity is available within one epoch (b).}
\label{fig:multiplot_vppl_training_error}
\end{figure}

Our results in Fig.~\ref{fig:convergence_phase_L}~and~\ref{fig:early_high_error} indicated that good NCE results could be attributed to its high error, i.e., high perplexity in the early stages of learning which may allow for broad exploration. We tried to enforce similar behaviour in softmax using a large learning rate in the search period. Figure~\ref{fig:multiplot_vppl_training_error}(a) presents the validation perplexity in the log scale for a large model with softmax (Y axis) as a function of a training epoch (X axis) and the learning rate (LR) which was increased to 1, 2, and 3 during search time. This arrangement increased the validation perplexity for the first few epochs, but the asymptotic convergence of softmax was not improved. When, in Fig.~\ref{fig:multiplot_vppl_training_error}(b), we compare the increased initial softmax perplexity with NCE perplexity during the progression of the first epoch, we can see that NCE has much larger perplexity at this stage even tough its learning rate is not larger than one. % the NCENLM training PPL is still much higher in the initial training phase compared to the softmax-based LM when learning rate is 1 and 3.
It might be a distinct characteristic of the NCE that helps to converge to a {\bf better local optimum} due to the initial high training error.

The numerical entries in Tab.~\ref{tab:weight_init_nce}, i.e., in all cells in the bottom right part of the table, contain both the intervals $U$ used to sample initial weights and the resulting perplexity (PPL) on a corresponding model. The results on the large model show that weight initialisation with lower variance led to better results, where the best perplexity of 69.995 was the best result that NCE achieved in our experiments. 

%% file: conclusion.tex
Language modelling techniques can use Noise Contrastive Estimation (NCE) to deal with the partition function problem during learning. Although it was known that NCE can outperform softmax (which computes the exact partition function) on large problems which are too big for softmax, its performance has never been shown to outperform softmax or other methods on tasks on which softmax is feasible and works well. In this paper, we showed that NCE can beat all the previously best results in the class of single-model methods achieving perplexity of 69.995. Our result establishes a new standard on the Penn Tree Bank dataset reducing the perplexity of the best existing method in this class by 8.405.